\documentclass[11pt]{article}
\usepackage[utf8]{inputenc}
\usepackage[T1]{fontenc}
\usepackage{amsmath,amssymb,amsthm}
\usepackage{booktabs}
\usepackage{graphicx}
\usepackage{hyperref}
\usepackage[margin=1in]{geometry}
\usepackage{xcolor}

\newtheorem{proposition}{Proposition}
\newtheorem{definition}{Definition}

\title{\textbf{When Is Emergent Consensus Real?}\\
A Measured Coupling Gain and a Validity Diagnostic for LLM Agent Societies}
\author{Dongxu Yang\thanks{Correspondence: \texttt{wayland0916@gmail.com}} \\
  DeepLethe}
\date{Preprint --- \today}

\begin{document}
\maketitle

\begin{abstract}
LLM ``agent societies'' are largely studied through demonstrations that report emergent
consensus, cooperation, or polarization---with no measurable control parameter, no theory of
when each regime appears, and no test of whether an emergent outcome is a genuine social
dynamic or an artifact of the underlying model. We introduce the \emph{coupling gain}
$\gamma$, a per-agent quantity measured \emph{directly from an LLM} by counterfactually
perturbing a neighbour's stated opinion. We show: (i) $\gamma$ is stable and
model-distinguishing---across five frontier models it ranges $0.15$--$0.43$ ($n{=}20$ reps,
bootstrap 95\% CIs $\le 0.025$ wide), and is invariant to prompt paraphrase; a control finds $\gamma$ for a social
neighbour $\approx \gamma$ for an impersonal numeric anchor, so $\gamma$ is an
\emph{evidence-coupling} coefficient rather than a uniquely social one. (ii) The macro regime
is organised by classical dynamics with \emph{measured} (not assumed) coefficients:
Friedkin--Johnsen for the consensus/pluralism axis, and the signed-Laplacian / structural
balance criterion for polarization. (iii) On the opinion-update tasks tested, frontier LLMs do
\emph{not} spontaneously backfire ($\beta\le 0$): default societies do not spontaneously
polarize---polarization is always \emph{induced} (the active-polarization $\beta>0$ branch
appears only in the FJ surrogate, not in the LLM agents). (iv) A randomized-initial-condition \emph{validity diagnostic}---the
(slope, bias) of final vs.\ initial opinion---separates genuine averaging from model-prior
artifacts; we rule out a boundary-censoring confound by construction using interior-valued
numeric facts, and \emph{apply the diagnostic to a published ``emergent consensus''
result} (Chuang et al.\ 2023), showing it \emph{conflates two mechanisms}---genuine averaging on
debatable claims with a model-prior artifact on settled facts---and that the effect is
model-specific.
(v) Probing transfer, coupling is \emph{context-dependent}: the pairwise $\gamma$ does not
predict multi-neighbour society outcomes---it can order them \emph{backwards}---whereas a
modality-matched \emph{group} coupling does (across sixteen closed-and-open models it predicts a
society's convergence with Pearson $r{=}{-}0.70$, permutation $p{=}0.008$). The coefficient that enters the regime laws of (ii)
is therefore this matched coupling, not the single-neighbour $\gamma$ of (i); emergent consensus
must be read from coupling measured in the target interaction, not a nominal influenceability.
Our contribution is a measurement protocol and a validity instrument, not new theory.
\end{abstract}

\section{Introduction}
Building a society of LLM agents and observing ``emergent'' behaviour has become a popular
methodology, but the dominant works are systems/demos: they report phenomena (a rumour
spreads, agents form cliques, opinions converge) without (a) a measurable parameter that
controls the outcome, (b) a theory predicting which regime appears, or (c) a test of whether
the phenomenon is a genuine social dynamic or an artifact of model sycophancy/homogenization.
We supply the missing rigour. Our unit of analysis is a single measured quantity, the
\emph{coupling gain} $\gamma$: how strongly an agent moves its stated opinion toward a
neighbour's. From a measured coupling and the influence network we
obtain a falsifiable regime prediction---though, as we show, the coupling that governs a
multi-neighbour society must itself be measured in the matched interaction---and from a
perturbation test, a lie-detector for emergent consensus.

\paragraph{Contributions (measurement-first, with a falsification test for aggregation).}
\begin{enumerate}
\item A \textbf{measured coupling gain $\gamma$} via counterfactual perturbation---the first
per-agent susceptibility coefficient measured directly on LLMs; stable, model-distinguishing,
paraphrase-invariant, and (by control) an evidence-coupling rather than uniquely-social
quantity.
\item A \textbf{(slope, bias) authenticity diagnostic} separating genuine averaging from
model-prior artifacts, with closed form $\kappa=\eta/(1-\gamma)$, applied to \textbf{re-analyse}
a published result (Chuang et al.\ 2023), showing it conflates two mechanisms.
\item A \textbf{negative result}: no frontier LLM spontaneously backfires ($\beta\le 0$), so
default societies cannot spontaneously polarize.
\item A \textbf{transfer finding with a falsification test}: the measurement transfers to
natural-language and vector-rating tasks, but coupling's \emph{value} does not---pairwise and
group susceptibility are \emph{negatively} associated (Spearman $\rho{=}{-}0.48$ over 15 models,
sign-robust to leave-one-model-out; suggestive at $p{=}0.07$), and only a modality-matched group
coupling predicts multi-neighbour outcomes. We prove (Prop.~4) that any \emph{additive}
aggregation forbids such a negative association, so the data are \emph{directionally inconsistent}
with the DeGroot/FJ/bounded-confidence family (consensus-conditional aggregation is one consistent
explanation).
\end{enumerate}
All code and per-run logs---including the contaminated-vs-clean convergence logs and the audit
script---are released.\footnote{\url{https://github.com/deeplethe/llm-coupling-gain}}
The regime predictions (Props.~1--3) are \emph{applications} of Friedkin--Johnsen
\cite{fj1990} and the signed-Laplacian / structural-balance criterion \cite{altafini2013},
with $\gamma,\beta$ \emph{measured} on the LLM rather than assumed.

\section{Related Work}
\textbf{LLM social simulation} (Generative Agents \cite{park2023}, Concordia
\cite{concordia2023}, AgentSociety \cite{agentsociety2025}) are systems/platforms with no
control/order parameter, no regime theorem, no spectral link; we add the quantitative layer.
\textbf{Emergent conventions} (Ashery, Aiello \& Baronchelli \cite{ashery2025}) show a genuine
disorder$\to$order transition but with random pairwise mixing---no network, no spectrum; they
defer network embedding. \textbf{Validity critiques} (PIMMUR \cite{pimmur2025};
Barrie \& T\"ornberg \cite{barrie2025}) argue much reported emergence is artifact via
qualitative checklists / data-leakage arguments; we provide the quantitative instrument.
\textbf{Classical opinion dynamics} (DeGroot \cite{degroot1974}; Friedkin--Johnsen
\cite{fj1990}; bounded confidence) establish that the spectrum of the influence matrix governs
consensus vs.\ disagreement; our novelty is that $\gamma$ is \emph{measured} on the LLM, and
the regime is a falsifiable prediction. (We verified that prior LLM opinion-dynamics papers
use LLM-generated updates with no measured gain and no spectral analysis.)
Single-agent self-conditioning \cite{selfcond2025} is the $N{=}1$ analogue.
\textbf{Concurrent LLM opinion-dynamics work} studies related effects without our
measured-gain + validity combination: \emph{imposed} self-vs-social weighting under varying
topology \cite{conformity2026} (a design parameter, not a counterfactually-\emph{measured}
$\gamma$); drivers of LLM social conformity \cite{confdrivers2026}; LLMs achieving structural
\emph{social balance} given explicit signed interactions \cite{socialbalance2024} (we instead
ask whether opinion-update agents \emph{spontaneously} form repulsive coupling, and find they
do not); validity via persona temporal stability \cite{stablepersona2026} and operational
replication \cite{opval2026}; and foundational principles of LLM opinion dynamics
\cite{principles2024}. None measures a per-agent coupling gain by counterfactual perturbation,
nor supplies an init-invariance authenticity test.

\section{The Coupling Gain}
\begin{definition}[Society as a coupled map]
$N$ agents on a row-stochastic influence matrix $W$; agent $i$ holds opinion
$x_i\in[0,100]$. One round: $x_i^{t+1}=F_i(x_i^t,\{x_j^t: W_{ij}>0\})$, $F_i$ the LLM
(black box; no update rule assumed).
\end{definition}
\begin{definition}[Coupling gain]
$\gamma_i$ is the sensitivity of agent $i$'s updated opinion to its neighbours' stated
opinions: present neighbour opinion $v$ across a grid, regress the updated opinion $u(v)$ on
$v$; $\gamma_i$ is the slope. $\gamma{=}0$: ignores neighbours; $\gamma{=}1$: fully adopts.
\end{definition}
Linearising $F$ about a fixed point gives the Friedkin--Johnsen (FJ) form
$x^{t+1}=\gamma W x^t + (1-\gamma)x^0$; $\gamma$ is the measured FJ susceptibility.

\section{Theory (applications of classical dynamics)}
\begin{proposition}[Consensus vs.\ pluralism]
For FJ dynamics with scalar $\gamma\in[0,1)$ and connected aperiodic row-stochastic $W$, the
stationary profile is $x^*=(1-\gamma)(I-\gamma W)^{-1}x^0$, a convex average of $x^0$, reached
at rate $\gamma$. As $\gamma\to 0$, $x^*\to x^0$ (full diversity); as $\gamma\to 1$, $x^*\to$
consensus at the Perron-weighted average.
\end{proposition}
\noindent\emph{Scope of Prop.~1 (anchoring matters).} Prop.~1 assumes persistent FJ anchoring
to the \emph{initial} opinion $x^0$. If agents instead anchor to their \emph{evolving} opinion
(pure DeGroot), every connected graph reaches consensus for \emph{any} $\gamma>0$; the
``$\gamma$ sustains pluralism'' claim then holds only on sparse/disconnected networks or when
the prompt induces self-anchoring (which our stubborn-agent runs supply, Sec.~\ref{sec:exp}).
We measure $\gamma$ under the FJ protocol and report this dependence rather than assuming it
away (App.~D).
\begin{proposition}[Authenticity diagnostic]
With an exogenous prior attractor $p$ of pull $\eta$ (doubly-stochastic $W$), the consensus
mean is $m^*=(1-\kappa)m_0+\kappa p$ with $\kappa=\eta/(1-\gamma)$. Hence regressing final
mean on initial mean gives $\mathrm{slope}=1-\kappa$ and $\mathrm{bias}=\kappa(p-m_0)$:
$\kappa\!\approx\!0$ (slope $\approx 1$, bias $\approx 0$) is genuine averaging (REAL);
$\kappa\!\to\!1$ (slope $\to 0$) is the model's prior imposed regardless of the group
(ARTIFACT).
\end{proposition}
\begin{proposition}[Polarization threshold]
Write signed-coupling dynamics $x^{t+1}=(I-\gamma L_s)x$ with signed Laplacian $L_s$. The
society polarizes iff $L_s$ has a negative eigenvalue (the destabilizing eigenvector is the
community/Fiedler vector). In a two-block mean field the community-difference mode grows by
$g=1+2\gamma\beta(a_{\mathrm{out}}/d)$: $\beta<0$ consensus, $\beta=0$ frozen (bounded
confidence), $\beta>0$ active polarization \emph{(this branch is a surrogate prediction)}.
Empirically $\beta\le 0$ for all measured models, so the $\beta>0$ branch is never realized on
agents---polarization is induced-only.
\end{proposition}
Proofs (Banach/Neumann for Prop.~1; mean-field + signed-Laplacian for Prop.~3, cf.\
\cite{altafini2013}) are in the appendix. Prop.~2's $\kappa$-identity passes a parameter-free
check (Sec.~\ref{sec:exp}). \emph{Which coupling enters these laws?} The $\gamma$ above is a
single scalar; Sec.~\ref{sec:exp} shows the \emph{single-neighbour} $\gamma$ is not
interchangeable with the \emph{multi-neighbour} coupling that actually drives a society, so the
regime laws must be fed the coupling measured in the matched interaction. We state this
carefully. A linearised multi-neighbour update with contraction rate $\rho<1$ still has a unique
stable consensus by Banach (reached at rate $\rho$); our data identify the \emph{ordering} of
$\rho$ across models with the group pull $p_{\mathrm{ft}}$---yielders contract, resisters do
not---and the held-out forecast (Sec.~\ref{sec:exp}) confirms exactly this ordering. We do
\emph{not} claim the convex-average / Perron-value refinements of Prop.~1 for the general
multi-neighbour map: those require the row-stochastic averaging structure and need not survive a
possibly anti-weighting update. The honest statement is ordinal: the coupling that controls
whether a society contracts is the matched group coupling, tracked by $p_{\mathrm{ft}}$, and the
single-neighbour $\gamma$ mis-estimates it (indeed orders societies backwards). The next
proposition explains \emph{why}.
\begin{proposition}[Additivity test: pairwise vs.\ group coupling]
Call a one-round update \emph{additively separable} if it depends on the neighbours only through
$S=\sum_j w_j\,\kappa(x_j)$ with weights $w_j\ge0$, $\kappa$ nondecreasing, and $G$ strictly
increasing in $S$ (i.e.\ $x^+=G(x,S)$; DeGroot, Friedkin--Johnsen, and bounded-confidence are all
of this form). Then the susceptibility
to a \emph{coherent group} of $m$ neighbours at value $v$ and to a \emph{single} neighbour at $v$
(own stance fixed) satisfy $p_{\mathrm{group}}=\big(\sum_{j\le m}w_j/w_1\big)\,\gamma_1\ge\gamma_1$
under sum-pooling, or $p_{\mathrm{group}}=\gamma_1$ under degree-normalised mean-pooling. Hence
$p_{\mathrm{group}}$ and $\gamma_1$ are sign-aligned with $p_{\mathrm{group}}\ge\gamma_1$: no
additively separable model can have $\gamma_1>0$ yet $p_{\mathrm{group}}\approx0$, nor a negative
$\gamma_1$--$p_{\mathrm{group}}$ correlation across agents.
\end{proposition}
\emph{The data lean against additivity for free-text deliberation (suggestively).} With \emph{numeric} neighbours the
group coupling amplifies the pairwise one ($\gamma_{\mathrm{grp}}{\approx}3\gamma$,
Fig.~\ref{fig:transfer}A)---additive-consistent. With \emph{free-text} neighbours, however, the
group pull $p_{\mathrm{ft}}$ is \emph{negatively} associated with $\gamma$ across models
(Spearman $\rho{=}{-}0.48$ over $15$ models, $-0.70$ on the $5$ closed; \emph{robust in sign to
leave-one-model-out}---negative dropping any single model, including DeepSeek, whose
$\gamma{=}0.43$ yet $p_{\mathrm{ft}}{\approx}0.01$). Under additivity Prop.~4 forbids any negative
association, so the data are \emph{directionally inconsistent} with additive aggregation---though
at this panel size the cross-model effect is \emph{suggestive} (permutation $p{=}0.07$), not a
significant falsification, and a larger panel is needed. Read together with the no-backfire
result (which excludes a repulsive kernel), this points to non-additive, consensus-conditional
aggregation---why the matched group coupling $p_{\mathrm{ft}}$, not the pairwise $\gamma$, tracks
the society's outcome and why $\gamma$ orders societies backwards.

\section{Experiments and Results}\label{sec:exp}
Setup: OpenRouter API; agents = LLMs; opinions $0$--$100$; controlled tasks. Models:
\texttt{deepseek-v4-pro}, \texttt{gpt-5.5}, \texttt{claude-opus-4.8},
\texttt{gemini-3.5-flash}, \texttt{qwen3.7-max}.

\subsection{The coupling gain: stable and model-distinguishing}
\begin{figure}[t]\centering
\includegraphics[width=0.62\linewidth]{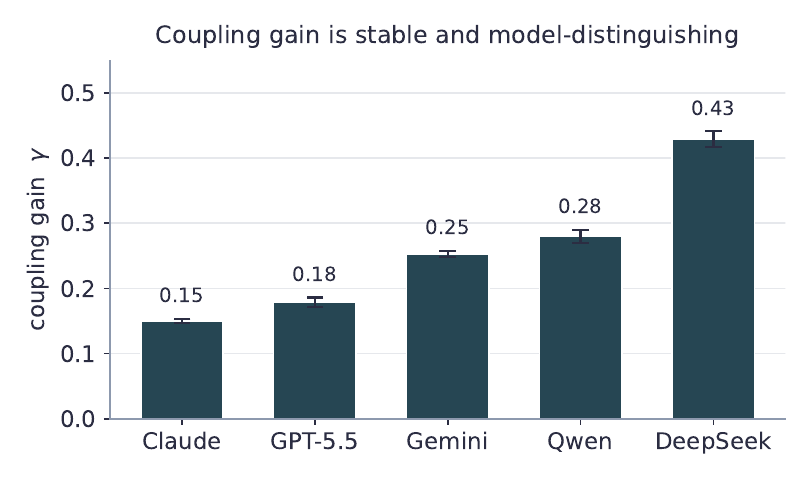}
\caption{Coupling gain $\gamma$ per model ($n{=}20$ reps, bootstrap 95\% CI).}
\label{fig:gamma}
\end{figure}

\begin{figure}[t]\centering
\includegraphics[width=0.6\linewidth]{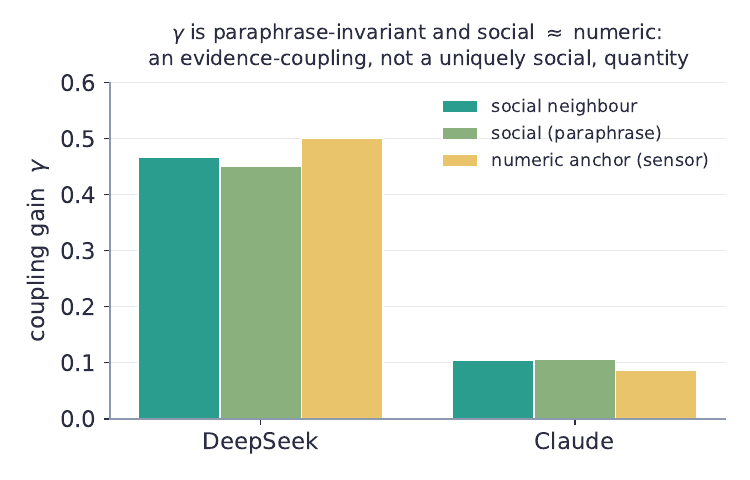}
\caption{Sycophancy control: $\gamma$ is paraphrase-invariant, and a social neighbour gives
nearly the same $\gamma$ as an impersonal numeric anchor---so $\gamma$ is an
\emph{evidence-coupling}, not a uniquely social, quantity.}
\label{fig:syco}
\end{figure}

\paragraph{Measured directly, with tight CIs (Fig.~\ref{fig:gamma}).}
$\gamma$ with bootstrap 95\% CI ($n{=}20$): DeepSeek $0.43\,[0.42,0.44]$, Qwen
$0.28\,[0.27,0.29]$, Gemini $0.25\,[0.25,0.26]$, GPT-5.5 $0.18\,[0.17,0.19]$, Claude
$0.15\,[0.15,0.15]$; CIs $\ll$ inter-model gaps.
$\gamma$ is invariant to prompt paraphrase, and $\gamma$(social neighbour) $\approx$
$\gamma$(numeric anchor) (DeepSeek $0.47$ vs $0.50$; Claude $0.11$ vs $0.09$;
Fig.~\ref{fig:syco}): an evidence-coupling, not uniquely social, quantity.

\subsection{Regimes: pluralism, consensus, and induced polarization}
\paragraph{No spontaneous backfire (negative result).}
With a strongly-opinionated agent facing a hostile neighbour, all five models move
\emph{toward} the neighbour or are inert ($\beta\le 0$) \emph{on opinion-update tasks}:
default societies do not spontaneously form repulsive coupling, so they cannot spontaneously
polarize (cf.\ \cite{socialbalance2024}, which studies balance given \emph{explicit} signed
interactions---a different setting). This is not a restatement of sycophancy training: sycophancy
concerns agreement with a \emph{single} interlocutor and is silent about \emph{cross-community}
dynamics---a model that mirrors whichever neighbour it faces could still amplify a community
split. That none does ($\beta\le0$) is an empirical fact about multi-agent structure, not a
corollary of the training objective.

\paragraph{Regime vs.\ $\gamma\times$density (P1).}
Real societies (final spread $\pm$std, $K{=}3$): on the sparse ring, stubborn Claude
($\gamma{=}0.15$) preserves pluralism ($6.9\pm1.9$) while DeepSeek ($\gamma{=}0.43$)
consensuses ($4.9\pm0.9$); on the complete graph both consensus (Claude $0.2$, DeepSeek
$3.6$). Matches Prop.~1.

\paragraph{Induced polarization.}
On a two-community SBM initialised low/high, DEFAULT agents converge while
confirmation-bias agents freeze, across all five models: gap-reduction ratio DEFAULT
$0.38$--$0.55$ vs.\ CONFIRM $1.02$--$1.09$ (disjoint bands). Formal test: paired across
models, $\text{CONFIRM}-\text{DEFAULT}=0.58\pm0.06$, $t(4){=}23.5$, $p<10^{-4}$.

\begin{figure}[t]\centering
\includegraphics[width=0.62\linewidth]{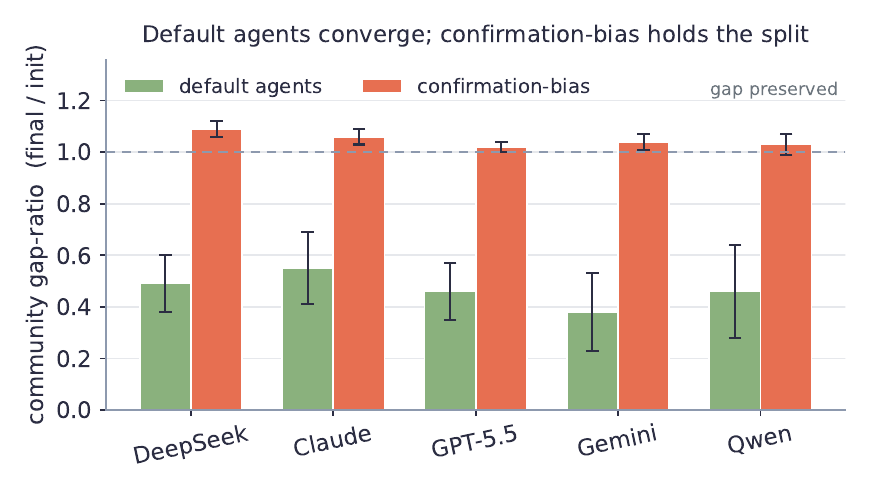}
\caption{Default agents converge across communities; confirmation-bias agents freeze
(gap-ratio $\to 1$). Polarization is induced, not spontaneous.}
\label{fig:induced}
\end{figure}

\subsection{A validity diagnostic and a re-analysis of Chuang et al.}
\begin{figure}[t]\centering
\includegraphics[width=0.66\linewidth]{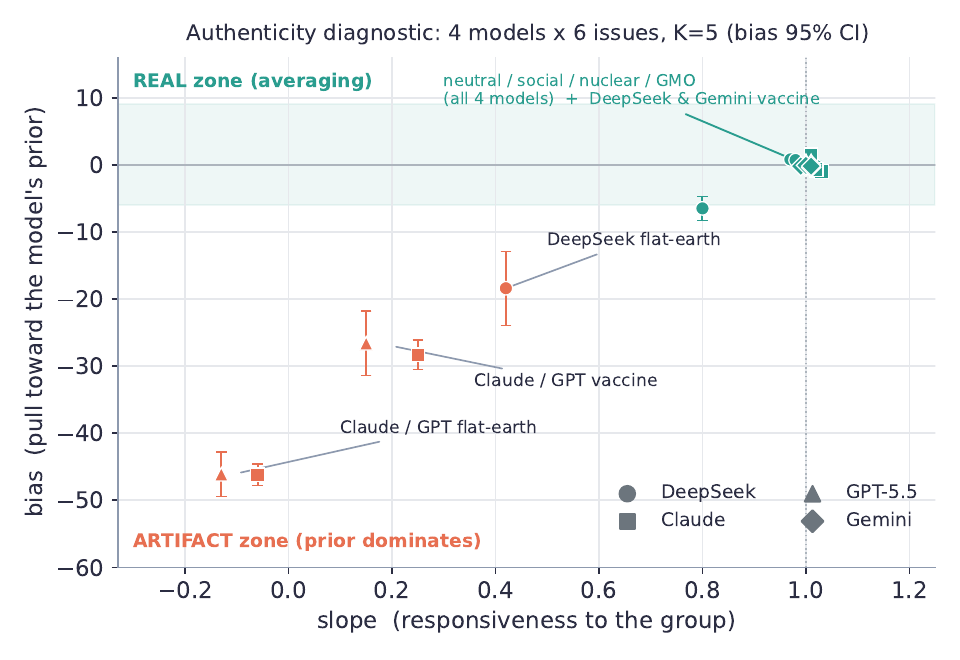}
\caption{Authenticity diagnostic, 4 models $\times$ 6 issues, $K{=}5$ (bias 95\% CI bars).
Debatable claims cluster at REAL (slope$\approx$1, bias$\approx$0); settled-fact claims are
prior-dominated ARTIFACTs for Claude/GPT, flat-earth only for DeepSeek, never for Gemini.}
\label{fig:auth}
\end{figure}

\paragraph{The diagnostic matrix (Table~\ref{tab:debunk}, Fig.~\ref{fig:auth}).}
We run the diagnostic over 6 issues $\times$ 4 models at $K{=}5$ with bias 95\% CIs. All four
models average faithfully on debatable claims (slope $\approx 1$, bias $\approx 0$); the
consensus becomes prior-dominated only on settled-fact claims, and the pattern is
\emph{model-dependent}: Claude and GPT-5.5 produce artifacts on both vaccine and flat-earth
(bias $-27$ to $-46$, CIs excluding $0$), DeepSeek only on flat-earth (it averages vaccine),
and \textbf{Gemini averages everything} (the diagnostic correctly reports no artifact). This
corrects Chuang et al.'s~\cite{chuang2023} blanket ``inherent accuracy bias $\to$ consensus'': it conflates
genuine averaging (debatable claims) with a model-prior artifact (settled facts), and the
effect is model-specific, not a universal property.

\begin{table}[t]\centering
\caption{Diagnostic battery: prior-pull \emph{bias} (mean $\pm$ 95\% CI, $K{=}5$) for 4 models.
REAL $=$ bias $\approx 0$ (genuine averaging); \textbf{bold} $=$ ARTIFACT (CI excludes $0$,
prior overrides the group). Artifacts concentrate on settled-fact claims and are model-specific.}
\label{tab:debunk}
\small
\begin{tabular}{lcccc}
\toprule
issue & DeepSeek & Claude & GPT-5.5 & Gemini\\
\midrule
remote work & $+0.8{\pm}0.6$ & $-0.9{\pm}0.3$ & $+0.6{\pm}0.4$ & $-0.3{\pm}0.3$\\
social media & $-0.2{\pm}0.8$ & $-1.0{\pm}0.6$ & $+0.3{\pm}0.2$ & $-0.0{\pm}0.2$\\
nuclear power & $+0.7{\pm}0.7$ & $-0.7{\pm}0.1$ & $+0.4{\pm}0.3$ & $-0.2{\pm}0.2$\\
GMO safety & $+1.0{\pm}0.6$ & $+1.5{\pm}0.4$ & $+0.6{\pm}0.3$ & $-0.1{\pm}0.2$\\
vaccines harmful & $-6.5{\pm}1.8$ & $\mathbf{-28.3{\pm}2.2}$ & $\mathbf{-26.6{\pm}4.8}$ & $-0.4{\pm}0.2$\\
Earth is flat & $\mathbf{-18.4{\pm}5.5}$ & $\mathbf{-46.2{\pm}1.6}$ & $\mathbf{-46.1{\pm}3.3}$ & $-0.2{\pm}0.3$\\
\bottomrule
\end{tabular}
\end{table}

\paragraph{Censoring ruled out by construction (interior facts, Table~\ref{tab:interior}).}
On opinion claims, the prior-attractor and $[0,100]$ censoring are confounded (the artifact
sits at the boundary, $p\approx 0$). We break the confound with numeric-fact claims whose
answer is \emph{interior} ($\approx71/60/21$): a society started at init$=15$ is pulled
\emph{up} to the interior value and one at init$=90$ down to it---convergence from both sides,
impossible under floor-censoring. The per-cell slope of final-on-init (slope $\approx 0$:
init-invariant attractor; $\approx 1$: averaging) is model-dependent: Qwen is a clean
attractor on all three facts, GPT-5.5/Claude on two, DeepSeek and Gemini average. The
diagnostic detects genuine prior-domination when present and correctly reports its absence
otherwise.

\begin{table}[t]\centering
\caption{Interior-fact init-slope (slope of final-mean on init-mean, mean $\pm$ std, $K{=}3$;
$p$ = independently solo-elicited prior). \textbf{Bold} $=$ slope$\approx$0 with final$\to$
interior $p$: a genuine attractor (censoring-immune, since the pull is \emph{toward an
interior value from both sides}). Model-dependent.}
\label{tab:interior}
\small
\begin{tabular}{lccc}
\toprule
model & earth-water ($p{=}71$) & body-water ($p{=}60$) & air-oxygen ($p{=}21$)\\
\midrule
Qwen & $\mathbf{0.00{\pm}0.00}$ & $\mathbf{0.01{\pm}0.02}$ & $\mathbf{0.00{\pm}0.00}$\\
GPT-5.5 & $\mathbf{0.10{\pm}0.18}$ & $0.72{\pm}0.36$ & $\mathbf{0.00{\pm}0.00}$\\
Claude & $\mathbf{0.07{\pm}0.03}$ & $0.72{\pm}0.22$ & $\mathbf{0.00{\pm}0.00}$\\
DeepSeek & $1.01{\pm}0.03$ & $0.95{\pm}0.04$ & $0.34{\pm}0.31$\\
Gemini & $1.01{\pm}0.03$ & $0.92{\pm}0.16$ & $0.89{\pm}0.01$\\
\bottomrule
\end{tabular}
\end{table}

\begin{figure}[t]\centering
\includegraphics[width=0.6\linewidth]{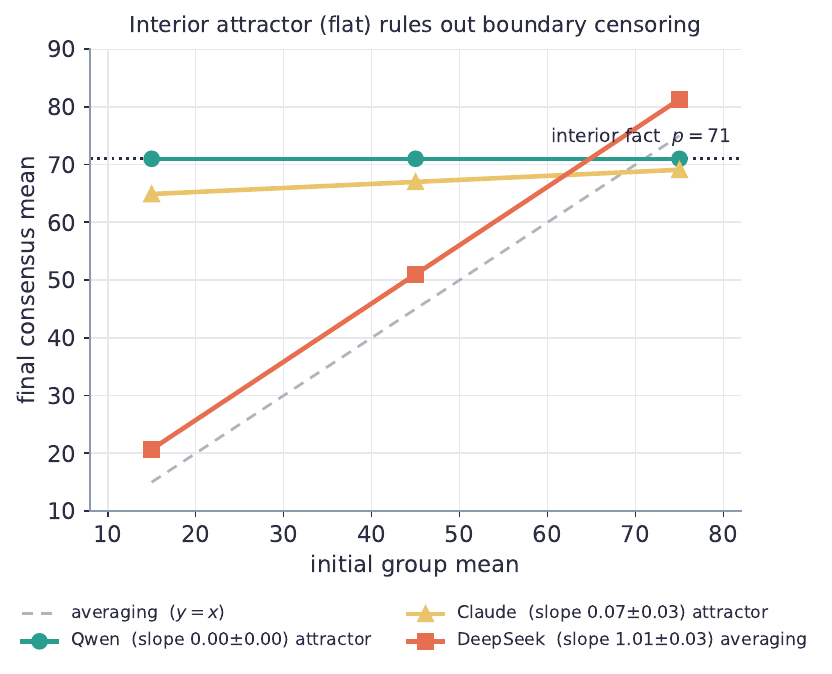}
\caption{Interior-fact convergence (Earth-water, $p{=}71$). A flat line (slope$\approx$0,
Qwen) is init-invariant convergence to the interior value---an upward pull from init$=15$
that floor-censoring cannot produce; the diagonal (DeepSeek) is averaging.}
\label{fig:interior}
\end{figure}

\subsection{Does the coupling gain transfer?}
\begin{figure}[t]\centering
\includegraphics[width=\linewidth]{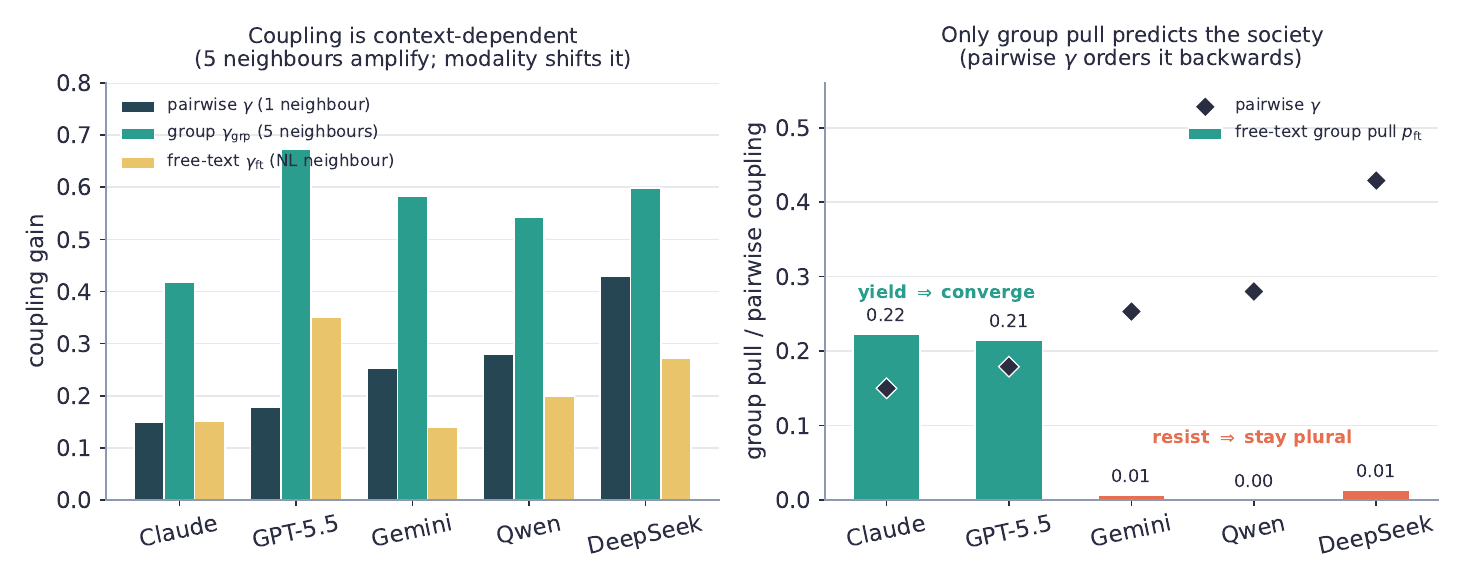}
\caption{\textbf{Coupling is context-dependent and only group coupling predicts the society.}
(A) A group of five neighbours amplifies coupling ${\sim}3\times$ over a single neighbour, and a
natural-language neighbour shifts it again (GPT up, DeepSeek down). (B) The free-text group pull
$p_{\mathrm{ft}}$ splits yielders (Claude/GPT) from resisters (DeepSeek/Gemini/Qwen); the
pairwise $\gamma$ (diamonds) orders them \emph{backwards}---DeepSeek has the highest pairwise
$\gamma$ yet the lowest group pull.}
\label{fig:transfer}
\end{figure}

\begin{figure}[t]\centering
\includegraphics[width=0.64\linewidth]{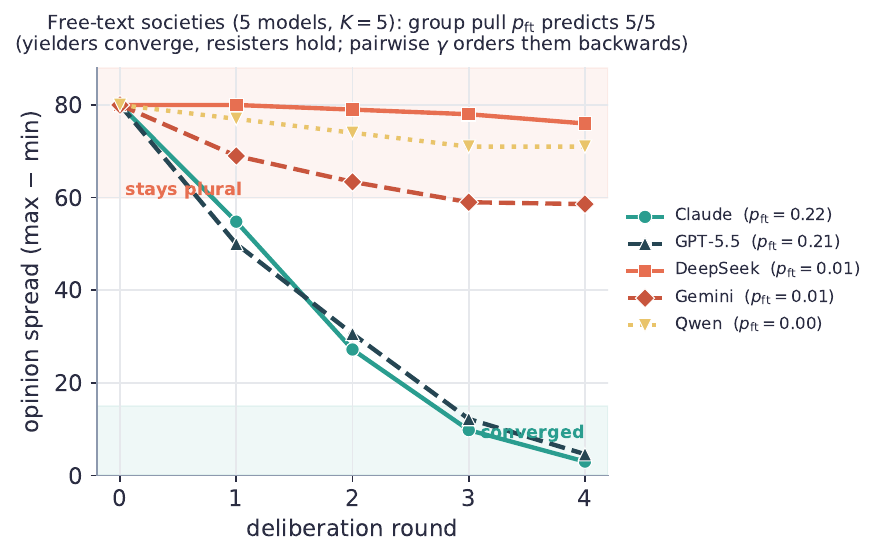}
\caption{Free-text six-agent societies ($K{=}5$, mean opinion spread per round). The two
high-group-pull models (Claude, GPT) converge; the three low-group-pull models (DeepSeek,
Gemini, Qwen) stay split---a \emph{held-out} $5/5$ match for $p_{\mathrm{ft}}$. DeepSeek has the
highest \emph{pairwise} $\gamma$ yet holds: the macro outcome tracks group, not pairwise,
coupling.}
\label{fig:traj}
\end{figure}

\paragraph{The measurement transfers; the value does not (Figs.~\ref{fig:transfer}--\ref{fig:traj}).}
We test whether the pairwise $\gamma$ predicts behaviour beyond the scalar single-neighbour
task. (i) \emph{Modality.} Measuring $\gamma$ with a natural-language neighbour \emph{argument}
instead of a number shifts its value: GPT-5.5 rises $0.18\to0.35$ while DeepSeek falls
$0.43\to0.27$ (Fig.~\ref{fig:transfer}A)---consistent with $\gamma$ being evidence-coupling (a
persuasive paragraph carries more evidence than a digit). (ii) \emph{Aggregation.} Facing a
\emph{group} of five neighbours rather than one amplifies coupling ${\sim}3\times$ for the
stubborn models ($\gamma_{\mathrm{grp}}$: Claude $0.15\to0.42$, GPT $0.18\to0.67$; $n{=}30$,
$R^2\ge0.92$). On a four-axis vector-rating task the measurement also transfers but adds little:
most models arithmetic-average each axis ($\gamma_d\approx0.5$ at $R^2\approx1$; Claude the
exception, $0.12$--$0.44$). (iii) \emph{The macro outcome is predicted by group, not pairwise,
coupling---a held-out test.} We measure $p_{\mathrm{ft}}$ (the susceptibility of an agent's
stance to a free-text group centred away from it) once, then run six-agent free-text town-halls
($K{=}5$) on all five models as a \emph{held-out} forecast. $p_{\mathrm{ft}}$ predicts the
convergence outcome \textbf{5/5} on the complete graph (upgraded below to a powered $n{=}16$
correlation across closed$+$open models, $r{=}{-}0.70$, $p{=}0.008$): the high-$p_{\mathrm{ft}}$ yielders (Claude $0.22$,
GPT $0.21$) converge (mean final spread $3.0$, $4.6$) while the low-$p_{\mathrm{ft}}$ resisters
(DeepSeek, Gemini, Qwen; $\le0.01$) stay plural (mean spread $59$--$76$;
Figs.~\ref{fig:transfer}B,~\ref{fig:traj}). The pairwise $\gamma$ orders the societies
\emph{backwards}---DeepSeek has the \emph{highest} pairwise $\gamma$ yet holds, while the two
\emph{lowest}-$\gamma$ models (Claude, GPT) converge---and pairwise $\gamma$ and $p_{\mathrm{ft}}$ are
negatively associated (Spearman $\rho={-}0.70$ on these five; $-0.48$ across all 15 models with
both measured, sign-robust to leave-one-model-out but only suggestive, $p{=}0.07$; cf.\ Prop.~4).
\emph{External validity (five conditions).} We re-run the forecast across a connectivity sweep
(ring deg-$2$, ring deg-$4$, complete), a larger society ($N{=}10$), and an asymmetric
initialisation. The predicted \emph{2-vs-3 separation} holds in every condition: the two
high-$p_{\mathrm{ft}}$ models end below all three low-$p_{\mathrm{ft}}$ models (ring deg-$2$:
$35,36$ vs $64$--$72$; ring deg-$4$: $8.6,7.6$ vs $57$--$75$; complete/asym: $2.0,7.4$ vs
$41$--$68$; $N{=}10$: $4.4,11$ vs $66$--$76$). Absolute convergence is
\emph{connectivity-dependent}: yielders' mean final spread falls with degree (ring ${\sim}35$,
ring2 ${\sim}8$, complete ${\le}5$) while resisters stay ${\sim}57$--$80$, so the sparse ring's
partial convergence is a mixing-rate effect, not a predictor failure. The ordering is robust to
topology, density, size, and initialisation. \emph{Data hygiene:} a failed API call is detected
(retried, then the whole run discarded), never silently recorded as a held opinion; all reported
numbers are over complete runs only (the same safeguard is applied in the surrogate-society code). \textbf{Takeaway:} the counterfactual \emph{measurement}
transfers to every setting, but coupling is context-dependent---pairwise and group
susceptibility can even anti-correlate---so society-level behaviour must be read from coupling
measured in the \emph{matched} interaction, not from a nominal one-on-one influenceability. This
sharpens the validity message rather than weakening it.

\paragraph{Reproducibility on open-weight models, and an independent prior check.}
Because closed frontier APIs drift over time, we replicate the protocol on \emph{eleven}
open-weight models (public weights, no drift: Llama-3.1-8B/70B, Llama-3.3-70B, Llama-4-Maverick,
Qwen-2.5-7B/72B, Mistral-Large/Small, Mixtral-8x22B, Gemma-2-27B, DeepSeek-Chat). Pairwise $\gamma$ is
again stable and model-distinguishing ($0.25$--$0.34$; bootstrap 95\% CIs $\le0.025$). The open
set \emph{spans the coupling range}---yielders (Qwen-2.5-7B $p_{\mathrm{ft}}{=}0.51$,
Mistral-Large $0.46$) and resisters (Llama-3.1-70B $-0.10$)---so the full yielder/resister
separation reproduces on open weights, and the $p_{\mathrm{ft}}\!\to\!$convergence relation
becomes a \emph{quantitative, powered} law rather than a five-model binary: across all
\textbf{sixteen} models (closed $+$ open), $p_{\mathrm{ft}}$ predicts the society's final spread
with Pearson $r{=}{-}0.70$ ($95\%$ bootstrap CI $[-0.87,-0.48]$; $r{=}{-}0.75$ dropping the lone
outlier), Spearman $\rho{=}{-}0.66$ (permutation $p{=}0.008$). The relation is
noisier on the diverse open set---one outlier (Mixtral-8x22B, high $p_{\mathrm{ft}}$ yet holds)
shows $p_{\mathrm{ft}}$ is a significant but imperfect predictor across architectures. Separately, to
check that Prop.~2's parameter-free test is not circular, we re-elicit the prior $p$ through an
\emph{independent} neutral factual-rating channel (no discussion framing): it matches the
diagnostic's $p$ (flat-earth ${\approx}0$, vaccines ${\approx}0$--$2$, and the ground-truth
control ``$71\%$ water'' ${\approx}100$ for the true statement), so $p$ is identified
independently of the dynamics it is used to explain.

\section{Limitations}
Small societies ($N{=}6$--$10$), few rounds; the free-text convergence (Fig.~\ref{fig:traj}) is
shown on five models at $K{=}5$; the predictor $p_{\mathrm{ft}}$ (measured on all
five at $n{=}30$) forecasts the convergence split held-out $5/5$ on the complete graph and
preserves the yielder/resister \emph{separation} across the external-validity
conditions (ring deg-$2$/deg-$4$/complete, $N{=}10$, asymmetric init), with absolute convergence
connectivity-dependent (slower on the sparse ring); still-larger societies, learned networks,
and non-opinion tasks remain future work; the propositions use a scalar-$\gamma$
linearisation whose limits we now demonstrate
empirically (Fig.~\ref{fig:transfer}); $\gamma$ is evidence-coupling, not uniquely social; the
active-polarization regime ($\beta>0$) is never observed on real agents (only on the FJ
surrogate); $\eta$ is identified phenomenologically and the opinion prior $p$ is solo-elicited (a
mild circularity in Prop.~2's parameter-free check that an independent $p$ calibration would
close---the interior-fact $p$ is ground truth and so immune). \emph{We are explicit about statistical scale}: the closed-model held-out forecast is $5/5$ on
five models, but the $p_{\mathrm{ft}}\!\to\!$convergence relation it rests on is now a
\emph{powered} correlation over sixteen closed$+$open models (permutation $p{=}0.008$); the
induced-polarization paired test, however, still has only four degrees of freedom
($t(4){=}23.5$), and societies remain small ($N{=}6$--$10$), so several claims rest on large,
non-overlapping effect sizes rather than large $n$. All numbers, per-rep
logs, and the contaminated-run audit are released.

\section{Conclusion}
On $0$--$100$ opinion-update tasks across five frontier models, a single measured
quantity---the coupling gain $\gamma$---plus a backfire coefficient $\beta$ organises the macro
outcome into consensus, pluralism, and (induced) polarization and predicts which appears; with
the (slope, bias) diagnostic (censoring-immune via interior facts) it separates genuine social
dynamics from model artifacts. We claim a reusable \emph{measurement protocol}, not a universal
law of agent societies: the active-polarization branch ($\beta>0$) is realized only in the FJ
surrogate (never measured on the agents), and pairwise $\gamma$ does \emph{not} transfer to
multi-neighbour, natural-language interaction---there the macro outcome is governed by the
modality-matched group coupling $p_{\mathrm{ft}}$ (Prop.~4), not the single-neighbour $\gamma$.
The aim is to replace demonstration with measurement.

\appendix

\section{Proof of Proposition 1 (consensus vs.\ pluralism)}
The map $T(x)=\gamma W x+(1-\gamma)x^0$ is affine with linear part $\gamma W$. In the
$\infty$-norm $\|\gamma W\|_\infty=\gamma\max_i\sum_j W_{ij}=\gamma<1$ (row-stochastic $W$), so
$T$ is a contraction; by Banach's theorem it has a unique fixed point with
$\|x^t-x^*\|_\infty\le\gamma^t\|x^0-x^*\|_\infty$ (rate $\gamma$). Solving $x^*=\gamma Wx^*+
(1-\gamma)x^0$ gives $x^*=(1-\gamma)(I-\gamma W)^{-1}x^0$ ($I-\gamma W$ invertible as
$\rho(\gamma W)=\gamma<1$). Expanding $(I-\gamma W)^{-1}=\sum_{k\ge0}(\gamma W)^k$, all terms
are entrywise $\ge0$ and the row sums of $(1-\gamma)\sum_k\gamma^kW^k$ equal
$(1-\gamma)\sum_k\gamma^k=1$, so each $x^*_i$ is a convex average of $x^0$. As $\gamma\to0$,
$x^*\to x^0$; as $\gamma\to1$, $(1-\gamma)(I-\gamma W)^{-1}\to\mathbf 1\pi^\top$ ($\pi$ the left
Perron vector), i.e.\ consensus at $\pi^\top x^0$. For reversible $W$ the stationary spread is
monotone decreasing in $\gamma$. \hfill$\square$

\section{Proof of Proposition 2 (authenticity slope and bias)}
With an exogenous attractor $p$ of pull $\eta$, $x^{t+1}=\gamma Wx^t+(1-\gamma-\eta)x^0+\eta
p\mathbf 1$ ($0\le\gamma+\eta<1$, $W$ doubly stochastic). The fixed point is
$x^*=(I-\gamma W)^{-1}[(1-\gamma-\eta)x^0+\eta p\mathbf1]$. Averaging and using that
doubly-stochastic $W$ preserves the mean, $\mathrm{avg}((I-\gamma W)^{-1}y)=\mathrm{avg}(y)/
(1-\gamma)$. With $m_0=\mathrm{avg}(x^0)$ and $\kappa=\eta/(1-\gamma)$,
$m^*=\frac{(1-\gamma-\eta)m_0+\eta p}{1-\gamma}=(1-\kappa)m_0+\kappa p$, so
$\mathrm{slope}=\mathrm dm^*/\mathrm dm_0=1-\kappa$ and $\mathrm{bias}=m^*-m_0=\kappa(p-m_0)$.
\hfill$\square$

\emph{Parameter-free check.} Claude on ``the Earth is flat'': slope $-0.07\Rightarrow
\kappa\approx1$, predicting bias $\approx p-m_0$; with solo-elicited $p\approx2$ and grid mean
$m_0\approx48$, $p-m_0\approx-46$, vs.\ measured $-46.2\pm1.6$.

\section{Proof sketch of Proposition 3 (polarization threshold)}
Write $x^{t+1}=(I-\gamma L_s)x$ with signed random-walk Laplacian $L_s=R-M$, $M_{ij}=
A_{ij}\sigma_{ij}/d_i$ ($\sigma=+1$ within, $-\beta$ across community), $R=\mathrm{diag}
(\sum_j M_{ij})$. An eigenpair $(\mu,v)$ of $L_s$ yields $T=I-\gamma L_s$ eigenvalue
$1-\gamma\mu$, which grows iff $\mu<0$: the society polarizes iff $L_s$ has a negative
eigenvalue, along its (community/Fiedler) eigenvector. In a balanced two-block mean field the
difference $\delta=m_A-m_B$ obeys $\delta^{t+1}=(1+2\gamma\beta\,a_{\mathrm{out}}/d)\,\delta$:
$\beta<0$ consensus, $\beta=0$ frozen, $\beta>0$ active polarization at rate
$\propto\beta\lambda_{\max}$ of the cross-community block. Empirically $\beta\le0$ for all five
models. \hfill$\square$

\section{Proof of Proposition 4 (additivity test)}
Write $x^+=G(x,S)$ with $S=\sum_j w_j\kappa(x_j)$, $w_j\ge0$, $G$ strictly increasing in $S$
($G_S>0$ on the relevant range: more neighbour agreement does not repel) and $\kappa$
nondecreasing ($\kappa'\ge0$). [The one-directional falsification used below needs only
$G_S\ge0$; the biconditional uses the strict $G_S>0$.]
Fix own stance $x=o$. A \emph{single} neighbour at $v$ gives $S_1=w_1\kappa(v)$ and
$\gamma_1=G_S(o,S_1)\,w_1\kappa'(v)$; a \emph{coherent group} of $m$ at $v$ gives
$S_m=\big(\sum_{j\le m}w_j\big)\kappa(v)$ and
$p_{\mathrm{group}}=G_S(o,S_m)\,\big(\sum_{j\le m}w_j\big)\kappa'(v)$. Each is a product of
nonnegative factors sharing $G_S\kappa'$, so $\gamma_1,p_{\mathrm{group}}\ge0$ are sign-aligned
and $p_{\mathrm{group}}=0\iff\gamma_1=0$; hence $\gamma_1>0\Rightarrow p_{\mathrm{group}}>0$. Under
degree-normalised mean-pooling the two signals coincide ($S_1=S_m=\kappa(v)$), giving exactly
$p_{\mathrm{group}}=\gamma_1$; under sum-pooling $p_{\mathrm{group}}/\gamma_1=
\big(\sum_{j\le m}w_j/w_1\big)\,G_S(o,S_m)/G_S(o,S_1)\ge1$ unless $G$ is strictly concave in $S$.
In every case the cross-agent map $\gamma_1\mapsto p_{\mathrm{group}}$ is nonnegative; a negative
correlation, or $\gamma_1>0$ with $p_{\mathrm{group}}\approx0$, is impossible. Contrapositive:
such an observation falsifies additive separability. \hfill$\square$

\section{Provenance and the boundary-censoring control}
Props.~1--3 apply classical results: Friedkin--Johnsen \cite{fj1990} / DeGroot
\cite{degroot1974} (consensus axis) and the signed-Laplacian / structural-balance criterion
\cite{altafini2013} (polarization); the novel atoms are the counterfactual measurement of
$\gamma,\beta$ on LLMs and the $\kappa$-identity. \emph{FJ vs.\ DeGroot (measured for the movers).} Regressing each agent's new stance on its
initial stance $x^0$, its previous stance $x^{t-1}$, and the neighbour mean: for the models that
actually move (the yielders Claude/GPT) the weight on $x^0$ is near zero ($|w_{x^0}|\le0.08$) and
that on $x^{t-1}$ dominates ($R^2\ge0.94$)---they anchor to their \emph{evolving} opinion
(DeGroot), not the initial one (FJ). The near-stationary resisters move too little to separate
$x^0$ from $x^{t-1}$ (the two regressors are collinear there), so their anchoring is
\emph{not identified}; but only the movers' anchoring bears on the pluralism caveat. Under
DeGroot a connected graph consensuses for any $\gamma>0$, so we restrict the ``$\gamma$ sustains
pluralism'' claim to sparse networks or prompt-induced self-anchoring---consistent with the
ring/dense split in Sec.~\ref{sec:exp}. \emph{Censoring:} on opinion claims the $\eta$-attractor and the
$[0,100]$ bound coincide at $p\approx0$; interior-valued numeric facts break the confound,
since an upward pull from $\mathrm{init}{=}15$ to an interior $p$ cannot be floor-censoring.

\section{Experimental details}
Agents are LLMs via OpenRouter (\texttt{deepseek-v4-pro}, \texttt{gpt-5.5},
\texttt{claude-opus-4.8}, \texttt{gemini-3.5-flash}, \texttt{qwen3.7-max}); opinions on
$0$--$100$; networks ring / ER / complete / two-community SBM. Replication: pairwise $\gamma$ at
$n{=}20$ (CIs are nonparametric bootstrap, $B{=}20{,}000$ resamples; script released); P1 and
induced at $K{=}3$ (robustness $N{=}16$, $K{=}5$); diagnostic battery $K{=}5$; flat-earth $K{=}10$;
interior facts $K{=}3$. Transfer: free-text and group coupling at $n{=}30$; free-text societies
($N{=}6$, also $N{=}10$; $T{=}4$) at $K{=}5$ on complete, ring (deg-$2$), and ring-2 (deg-$4$)
graphs and an asymmetric initialisation. A failed API call is retried, then the run is discarded
(never recorded as a held opinion); reported numbers are over complete runs only. All per-run
logs (and the contaminated-run audit) are released.

\end{document}